\pdfoutput=1

\documentclass[11pt]{article}

\usepackage[final]{acl}

\usepackage{times}
\usepackage{latexsym}
\usepackage{booktabs} 
\usepackage{multirow} 
\usepackage[T1]{fontenc}
\usepackage{algorithm}
\usepackage{algorithmic}
\usepackage[utf8]{inputenc}

\usepackage{microtype}

\usepackage{inconsolata}

\usepackage{graphicx}
\usepackage{CJKutf8}
\title{Developing and Utilizing a Large-Scale Cantonese Dataset for Multi-Tasking in Large Language Models}

\author{
\bf Jiyue Jiang$^{\heartsuit}$,
Alfred Kar Yin Truong$^{\spadesuit}$,
Yanyu Chen$^{\heartsuit}$,  
Qinghang Bao$^{\spadesuit}$,
Sheng Wang$^{\spadesuit}$, \\
\bf Pengan Chen$^{\spadesuit}$,
Jiuming Wang$^{\heartsuit}$,
Lingpeng Kong$^{\spadesuit}$, 
Yu Li$^{\heartsuit}$,
Chuan Wu$^{\spadesuit}$ \\
$^{\heartsuit}$ The Chinese University of Hong Kong, $^{\spadesuit}$ The University of Hong Kong \\
{\tt
\{jiangjy, jmwang\}@link.cuhk.edu.hk,
} \\
{\tt
alfred.truong@gmail.com,
redbaron.chen@connect.polyu.hk,
} \\
{\tt
\{bill6176, u3009618, cpa2001\}@connect.hku.hk,
} \\
{\tt
liyu@cse.cuhk.edu.hk,
\{lpk, cwu\}@cs.hku.hk
}
}
\vspace{5cm}

\begin{document}
\maketitle
\begin{abstract}
High-quality data resources play a crucial role in learning large language models (LLMs), particularly for low-resource languages like Cantonese. Despite having more than 85 million native speakers, Cantonese is still considered a low-resource language in the field of natural language processing (NLP) due to factors such as the dominance of Mandarin, lack of cohesion within the Cantonese-speaking community, diversity in character encoding and input methods, and the tendency of overseas Cantonese speakers to prefer using English. In addition, rich colloquial vocabulary of Cantonese, English loanwords, and code-switching characteristics add to the complexity of corpus collection and processing. To address these challenges, we collect Cantonese texts from a variety of sources, including open source corpora, Hong Kong-specific forums, Wikipedia, and Common Crawl data. We conduct rigorous data processing through language filtering, quality filtering, content filtering, and de-duplication steps, successfully constructing a high-quality Cantonese corpus of over 2 billion tokens for training large language models. We further refined the model through supervised fine-tuning (SFT) on curated Cantonese tasks, enhancing its ability to handle specific applications. Upon completion of the training, the model achieves state-of-the-art (SOTA) performance on four Cantonese benchmarks. After training on our dataset, the model also exhibits improved performance on other mainstream language tasks.

\end{abstract}

\section{Introduction}

High-quality data resources are essential for the advancement of large language models~\cite{jiang2024well}, particularly for languages with limited digital resources, such as Cantonese. Although Cantonese boasts over 85 million native speakers~\cite{xiang2024cantonese, jiang2024well}, predominantly located in southern China and among Chinese communities worldwide, it remains classified as a low-resource language within the domain of NLP. This is mainly due to the dominance of Mandarin, the lack of uniformity within the Cantonese-speaking community, and the diversity of character encoding and input methods. In addition, overseas Cantonese speakers tend to use English, which further hinders the development of Cantonese in the NLP domain.

\begin{figure*}[t]
    \centering
    \includegraphics[width=15cm]{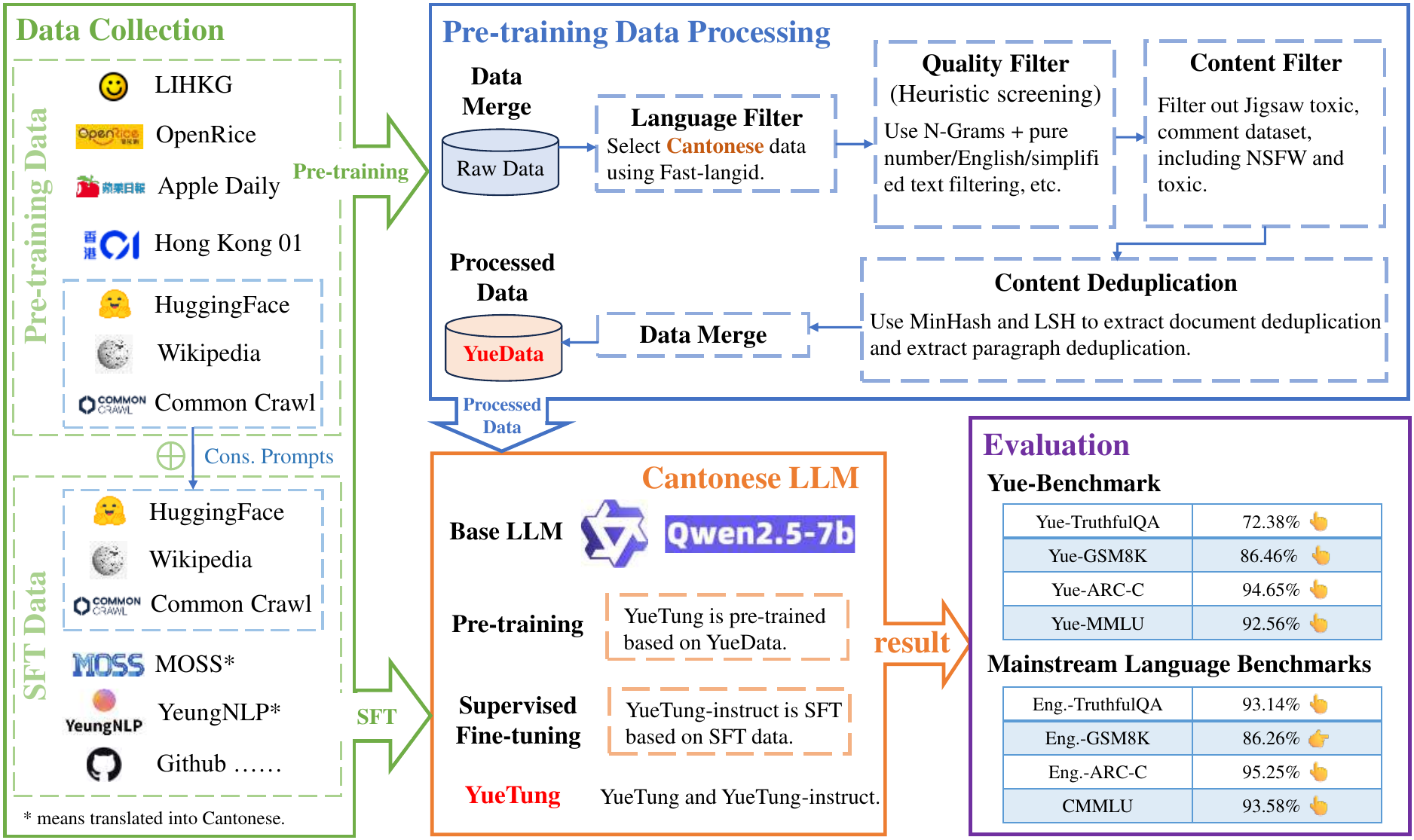}
    \caption{Overview of our work. We construct Cantonese pre-training and SFT data, apply language and quality filters to the former, and derive the latter from it. The base model of YueTung is the \textbf{Qwen-2.5-7b} model, which is trained on the YueData. YueTung achieve SOTA performance on Cantonese benchmarks, and its performance on mainstream language benchmarks not only did not decline but actually improved.}
    \label{fig:modeloverview}
\end{figure*}

The rich colloquial vocabulary of Cantonese, its English loanwords, and the widespread phenomenon of code-switching make corpus collection and processing more complex. Compared to Modern Standard Chinese, there is a significant disparity between spoken and written Cantonese; many colloquial expressions lack a standardized written form. Furthermore, Cantonese writing involves the conversion between traditional and simplified characters, as well as the use of unique Cantonese characters and words~\cite{yu2022automatic, xiang2024cantonese}. These factors increase the difficulty of text data normalization and processing. These challenges have led to a scarcity of high-quality Cantonese corpora, limiting the performance enhancement of LLMs in the Cantonese context.

To address these issues, we collect diverse Cantonese text data to construct a high-quality Cantonese dataset. Data sources include open-source corpora, Hong Kong-specific forums such as LIHKG\footnote{\url{https://lihkg.com/}}, OpenRice\footnote{\url{https://www.openrice.com/}}, the Cantonese version of Wikipedia\footnote{\url{https://dumps.wikimedia.org/zh_yuewiki/}}, and Common Crawl data\footnote{\url{https://commoncrawl.org/}}, etc. During the data collection process, we pay attention to the variations in Cantonese usage across different regions and platforms. We utilize custom web crawlers and data extraction tools to efficiently gather large amounts of text from these sources.

To ensure the quality and purity of the data, we establish and executed a stringent data processing workflow. First, we perform language filtering on the collected texts using language identification models to ensure only Cantonese content is retained. Next, we conduct quality filtering based on a series of heuristic rules to select high-quality texts. We also apply specialized classifiers to detect and remove harmful content, such as toxic language and sensitive information. In addition, techniques like MinHash and Locality Sensitive Hashing (LSH)~\cite{pauleve2010locality} are used for deduplication, ensuring the corpus's uniqueness and diversity. After pre-training the model on this extensive corpus, we apply SFT using additional Cantonese datasets to further enhance its performance on downstream tasks.

Through these efforts, we successfully build a high-quality Cantonese corpus containing over 2 billion tokens, laying a solid foundation for training large language models. In model performance evaluations, our model achieve industry-leading performance on four Cantonese benchmark tests, accurately handling Cantonese-specific vocabulary and expressions while generating fluent and natural text. Notably, after training with our data, the model also demonstrate performance improvements on other mainstream language tasks, proving that high-quality Cantonese data contributes to the overall performance enhancement of the model.


\begin{table*}[!ht]\small
    \centering
    \begin{tabular}{l|l|c|c|c|c|c|c}
        \toprule
        \multirow{2}{*}{\textbf{Source}} & \multirow{2}{*}{\textbf{Record type}} & \multicolumn{2}{c|}{\textbf{Total}} & \multicolumn{4}{c}{\textbf{Record Char Distribution}} \\
        \cmidrule{3-8}
        &  & \textbf{chars} & \textbf{records} & \textbf{min} & \textbf{median} & \textbf{mean} & \textbf{max} \\
        \midrule
        Wikipedia & page contents & 40,398,140 & 137,342 & 4 & 91 & 294 & 60,004 \\
        raptorkwok & cantonese\_sentences & 664,993,424 & 30,150,987 & 0 & 16 & 22 & 2,328 \\
        Apple Daily & html articles & 54,156,758 & 81,081 & 200 & 535 & 668 & 21,298 \\
        LIHKG (1-2.8m) & threads & 1,582,487,817 & 2,873,877 & 7 & 360 & 551 & 80,687 \\
        LIHKG (2.8m-3.8m) & sub-threads & 839,667,516 & 29,563,007 & 0 & 13 & 28 & 4,705 \\
        OpenRice & restaurant reviews & 490,181,056 & 1,234,262 & 0 & 315 & 397 & 477,672 \\
        \bottomrule
    \end{tabular}
    \caption{Custom scraped corpora, count of characters is language and punctuation agnostic (statistics are indicative).}
    \label{tab:used_corpora}
\end{table*}

\begin{table*}[!ht]\small
    \centering
    \begin{tabular}{l|l|l}
        \toprule
        \textbf{Corpus Name} & \textbf{Size} & \textbf{Source} \\
        \midrule
        CanCorp\footnote{\url{https://github.com/fcbond/hkcancor}}~\cite{Lee1998CancorpTH} & 1M tokens & child speech research \\
        HKCAC~\cite{hkcac} & 170K tokens & phone-in programs and radio \\
        HKCancor\footnote{\url{https://github.com/fcbond/hkcancor}}~\cite{Wong2015TheHK} & 230K tokens & speech and radio programs \\
        HKCC~\cite{chin2015linguistics} & 1M tokens & audio from 1940-1970 HK movies \\
        UD\_Cantonese-HK\footnote{\url{https://github.com/UniversalDependencies/UD_Chinese-HK}}~\cite{nivre-etal-2017-universal} & -- & film subtitles and LegCo proceedings \\
        MyCanCorp\footnote{\url{https://github.com/liesenf/MYCanCor}}~\cite{Liesenfeld18} & 20 hours of audio & Malaysian Cantonese speech \\
        Common Voice zh-HK\footnote{\url{https://commonvoice.mozilla.org/en/datasets}}~\cite{DBLP:journals/corr/abs-1912-06670} & 109 hours of audio & Mozilla audio collection program \\
        DRCD\footnote{\url{https://github.com/DRCKnowledgeTeam/DRCD}}~\cite{shao2019drcdchinesemachinereading} & 10K paragraphs & Wikipedia \\
        CantoMap\footnote{\url{https://github.com/gwinterstein/CantoMap}}~\cite{winterstein-etal-2020-cantomap} & 106K tokens & 12.8hrs of speech \\
        MDCC\footnote{\url{https://github.com/HLTCHKUST/cantonese-asr}}~\cite{yu-etal-2022-automatic} & 73.6 hours of audio & clean speech from audiobooks \\
        \bottomrule
    \end{tabular}
    \caption{Open-source corpora from previous studies.}
    \label{tab:open_source_corpora}
\end{table*}

\section{Large-Scale Cantonese Data: YueData}

\subsection{Pre-Trained Cantonese Data}

\subsubsection{Pre-Trained Data Collection}

Although there are many challenges 
in gathering Cantonese text, we build a large-scale corpus with a focus on spoken Cantonese, and proceed in two phases: corpus collection and post-processing. 

    

\paragraph{Corpus Collection}
\label{subsec:2_data_1_alfred}
Cantonese text is gathered from the following sources: (1) open-source corpora; (2) HK transcriptions; (3) HK online publications; (4) HK online forums; (5) Chinese entries from Common Crawl\footnote{\url{https://commoncrawl.org/}}. 

We focuse on leveraging existing open-source resources and subsequently scraping Cantonese data from Hong Kong-centric resources known to be of high quality and which predominantly employ less formal language, resembling spoken Cantonese. The emphasis on these resources is partly due to familiarity with the sources and the status of Hong Kong Cantonese as a de facto standard\footnote{HK Cantonese has great reach among Cantonese speaking communities as (1) HK has a relatively large and uniform speaker base, (2) emigration from HK seeded many overseas diasporas, and (3) HK was an early producer of Cantonese media (movies, TV dramas, and pop culture), thereby widely consumed and recognized.}. In addition, we interface with Common Crawl to amass a broader corpus of Chinese text.


\textbf{Open-source corpora:} (1) Wikipedia serves as a primary source due to its comprehensive data availability. The Wikipedia pages are systematically archived, categorized by language, and are accessible for batch downloading\footnote{\url{https://dumps.wikimedia.org/backup-index.html}}. Specifically, the Cantonese language content is designated as zh\_yuewiki\footnote{\url{https://dumps.wikimedia.org/zh_yuewiki/}}, from which the extraction of page contents is straightforward. (2) Prior research: We reviewed data utilized in existing studies on Cantonese linguistics and NLP as referenced in \autoref{tab:open_source_corpora}. These corpora, however, are typically limited in size, often comprising less than a million characters. Given our need for more extensive datasets, these were not included in our study. (3) Huggingface: This platform hosts numerous large-scale datasets\footnote{\url{https://huggingface.co/datasets}}, though the origins of these datasets are not always transparent. Noteworthy within the context of Cantonese language resources are several datasets, including raptorkwok/cantonese\_sentences\footnote{\url{https://huggingface.co/datasets/raptorkwok/cantonese_sentences}}, which includes approximately 30.2 million sentences likely sourced from educational materials featuring colloquial text. Another significant dataset is AlienKevin/LIHKG\footnote{\url{https://huggingface.co/datasets/AlienKevin/LIHKG}}, comprising around 2.8 million discussion threads extracted from LIHKG\footnote{\url{https://lihkg.com/}}, a popular Hong Kong forum akin to Reddit where users engage in informal discussions, frequently using vernacular and slang.
        
        
        
        

\begin{table*}[!ht]\small
    \centering
    \begin{tabular}{l|l|l}
        \toprule
        \textbf{Corpus Name} & \textbf{Size} & \textbf{Source} \\
        \midrule
        CanCorp\footnote{\url{https://github.com/fcbond/hkcancor}}~\cite{Lee1998CancorpTH} & 1M tokens & child speech research \\
        HKCAC~\cite{hkcac} & 170K tokens & phone-in programs and radio \\ %
        HKCancor\footnote{\url{https://github.com/fcbond/hkcancor}}~\cite{Wong2015TheHK} &  230K tokens & speech and radio programs \\ %
        HKCC~\cite{chin2015linguistics} & 1M tokens & audio from 1940-1970 HK movies \\ %
        UD\_Cantonese-HK\footnote{\url{https://github.com/UniversalDependencies/UD_Chinese-HK}}~\cite{nivre-etal-2017-universal} & -- & film subtitles and LegCo proceedings \\ %
        MyCanCorp\footnote{\url{https://github.com/liesenf/MYCanCor}}~\cite{Liesenfeld18} & 20 hoours of audio & Malaysian Cantonese speech \\
        Common Voice zh-HK\footnote{\url{https://commonvoice.mozilla.org/en/datasets}}~\cite{DBLP:journals/corr/abs-1912-06670} & 109 hours of audio & Mozilla audio collection program \\ 
        DRCD\footnote{\url{https://github.com/DRCKnowledgeTeam/DRCD}}~\cite{shao2019drcdchinesemachinereading} & 10K paragraphs & Wikipedia \\
        CantoMap\footnote{\url{https://github.com/gwinterstein/CantoMap}}~\cite{winterstein-etal-2020-cantomap} & 106K tokens & 12.8hrs of speech \\ 
        MDCC\footnote{\url{https://github.com/HLTCHKUST/cantonese-asr}}~\cite{yu-etal-2022-automatic} & 73.6 hours of audio & clean speech from audiobooks \\
        \bottomrule
    \end{tabular}
    \caption{Open-source corpora from previous studies.}
    \label{tab:open_source_corpora}
\end{table*}



\textbf{HK transcriptions: governmental bodies and TV/movie subtitles.} 
Transcriptions of RTHK radio programs and HK Legco discussions exist and are of good quality\footnote{Both are government organizations and are transcribed from spoken language, so are unlikely to be vulgar, overly formal, or overly informal}. Although we find some transcriptions, we cannot find large accessible repositories of them, and so we do not make use of this resource (it could prove fruitful for more resourceful researchers). Another potential source of spoken text includes TV and movie subtitles. One can either (1) \textbf{grab pre-generated files}: there are online forums that host \texttt{.srt} extension files that media players use to display closed captions, and these files are either created by original content creators or the open-source community; or (2) \textbf{generate closed captions with ASR tools}: open-source and paid ASR tools can extract text from audio while proprietary tools\footnote{Automatic transcription by Google of YouTube videos} can also be leveraged.

Although this appears to be a promising route, both sources present limitations. Large repositories of \texttt{.srt} files are scarce, and the fidelity of subtitles to spoken Cantonese varies greatly\footnote{Closed captions are often written in formal Chinese to convey meaning, as formal Chinese is more concise and easier to type and read; it is not always spoken Cantonese}. In addition, Cantonese ASR is an active research area, and ASR output would likely require validation before use. Therefore, we decide not to use transcriptions, as it is not the main purpose of this paper.

\textbf{HK online publications: Apple Daily.}
We focus on HK publications using less formal Cantonese: (1) \textbf{Apple Daily}, the now-defunct publication by Next Digital; (2) \textbf{HK01.com}\footnote{\url{HK01.com}}, an influential online portal covering popular news in HK.

We extract content only from Apple Daily using 120 web archives\footnote{\url{https://archive.fart.website/archivebot/viewer/job/201910102213472u3qb}, \url{https://archive.fart.website/archivebot/viewer/job/202008102032142u3qb}, and \url{https://archive.fart.website/archivebot/viewer/job/202106170425282u3qb}}. We do not scrape text from HK01.com, though it is a rich source of Cantonese text for NLP researchers.

\textbf{HK online forums: LIHKG, OpenRice.}
Online forums are excellent sources of informal Cantonese due to loose language enforcement, allowing users to discuss any topic freely. We focus on two forums\footnote{Other shortlisted forums not tackled include: \url{https://www.discuss.com.hk/} (GitHub repo \url{https://github.com/vanatteveldt/discusshk/blob/master/scrape_discusshk.py}), \url{https://m.hkgolden.com/}, \url{https://www.baby-kingdom.com}, and \url{https://www.babydiscuss.com/}}: (1) \textbf{LIHKG}\footnote{\url{https://lihkg.com/}}, a popular multi-category forum among HK youths; (2) \textbf{OpenRice}\footnote{\url{https://www.openrice.com/}}, a widely used restaurant review platform rich in spoken Cantonese.

We write custom web-scrapers~\cite{Truong_LIHKG_scraper_2024,Truong_openrice_scraper_2024}\footnote{Improved upon papatekken's (\url{https://github.com/papatekken/simple-LIHKG-scraper-with-python}) LIHKG scraper and francoishideyos's (\url{https://github.com/francoishideyos/openrice_recommendator}) OpenRice scraper} to extend LIHKG coverage from 2.8 million to 3.8 million threads and to collect numerous restaurant reviews from OpenRice.

\textbf{Chinese Entries from Common Crawl.}
We use Common Crawl to build our Chinese corpus\footnote{Chinese text represents only \href{https://commoncrawl.github.io/cc-crawl-statistics/plots/languages}{5\%} of recent Common Crawl indexes}. Using AWS Athena, we query CDX Index files and employ language identifiers\footnote{Language annotation was \href{https://commoncrawl.org/blog/august-2018-crawl-archive-now-available}{introduced} from \textbf{CC-MAIN-2018-39} onwards; we use language predictors where it was not \href{https://commoncrawl.org/errata/missing-language-classification}{provided}} to pinpoint records with Chinese text. We write a custom crawler~\cite{Truong_cc_cached_downloader_2024} to handle this task.

\subsubsection{Pre-Trained Data Processing}

\begin{table*}[h!]\small
\centering
\begin{tabular}{l|c|c|c|c|c|c}
\toprule
\multirow{2}{2.7cm}{\textbf{Models \\ (7-8b scale)}} & \multicolumn{3}{c|}{\textbf{0-shot}} & \multicolumn{3}{c}{\textbf{5-shot}} \\
\cmidrule{2-7}
& \textbf{Rouge-l} & \textbf{Bleu-4} & \textbf{BERTScore}  & \textbf{Rouge-l} & \textbf{Bleu-4} & \textbf{BERTScore}  \\
\midrule
Qwen-2.5-7b & 18.51 	& 	12.28  & 66.07  & 6.83 	& 	8.07  & 58.97\\
Llama-3.1-8b & 13.82 & 10.33 & 66.97 & 26.18 & 15.20 & 70.28 \\
Yi-1.5-6b & 1.21 	& 	4.60   & 42.15 & 1.04 	& 	6.15  & 53.85  \\
Internlm-2.5-7b-chat & 7.13 & 8.00 & 63.48 & 4.05 & 7.19 & 67.61 \\
\textbf{YueTung-7b} & \textbf{33.95} & \textbf{12.54} & \textbf{71.33} & \textbf{35.12} & \textbf{13.52} & \textbf{72.38} \\
\bottomrule
\toprule
\multirow{2}{2.7cm}{\textbf{Models \\ (> 7-8b scale)}} & \multicolumn{3}{c|}{\textbf{0-shot}} & \multicolumn{3}{c}{\textbf{5-shot}} \\
\cmidrule{2-7}
& \textbf{Rouge-l} & \textbf{Bleu-4} & \textbf{BERTScore}  & \textbf{Rouge-l} & \textbf{Bleu-4} & \textbf{BERTScore}  \\
\midrule
Qwen-2.5-72b & 13.03 	& 	9.64   & 66.94 & 20.23 	& 	12.87   & 69.53 \\
Mistral-large-2 & 19.72 & 13.01 & 69.06 & 31.38 & 18.61 & 72.07 \\
Llama-3.1-70b & 21.03 & 14.30 & 68.31 & 34.72 & \textbf{20.54} & 70.80 \\
Phi-3-medium & 18.70 & 12.00 & 67.36 & 22.00 & 13.72 & 67.57 \\
Gemma-2-27b & 8.09 & 8.44 & 64.41 & 11.33 & 9.98 & 63.66  \\
Yi-1.5-34b & 15.41 & 11.11 & 67.57 & 20.30 & 13.20 & 69.50 \\
Internlm-2.5-20b-chat & 6.96 & 7.73 & 62.99 & 3.28 & 6.06   & 66.99 \\
ERNIE-Turbo & 17.91 & 11.30 & 66.71  & 21.19 & 12.19 & 68.29 \\
Sensechat-5 & 24.75 & \textbf{15.11} & 68.43 & 32.45 & 19.70 & 70.02  \\
Claude-3.5 & 14.23 & 9.95 & 67.56 & 12.66 & 10.06 & 68.12  \\
GLM-4 & 13.44 & 10.07 & 67.26 & 23.57 & 14.28 & 70.30 \\
ChatGPT & 25.07 & 14.81 & 67.78 & 31.84 & 18.42 & 70.41 \\
GPT-4 & 19.47 & 13.45 & 68.99 & 28.43 & 16.74 & 71.26 \\
\textbf{YueTung-7b} & \textbf{33.95} & 12.54 & \textbf{71.33} & \textbf{35.12} & 13.52 & \textbf{72.38} \\
\bottomrule
\end{tabular}
\caption{Results of the comparison between texts generated by YueTung-7b and baselines in \textbf{Yue-TruthfulQA} based on 0-shot and 5-shot settings and the ground truth.}
\label{TruthfulQA_Cant}
\end{table*}




In our Cantonese data processing workflow, we follow the methodological framework established by Dolma~\cite{dolma}, adhering to the ``garbage in, garbage out" principle to ensure data integrity and quality. This process includes several stages specifically tailored for Cantonese: language filtering~\cite{fastlang2023}, heuristic-based quality filtering~\cite{rae2022gopher, 2019c4}, content filtering, and deduplication.

\textbf{Language Filtering.} We use the automatic language identification tool Fast-Langid~\cite{fastlang2023}, an extension of FastText~\cite{fasttext} capable of identifying Cantonese, to build our dataset. We exclude data sources already pre-filtered for Cantonese, like OpenRice, due to imperfections in language models~\cite{multilang}. We proceed to process documents tagged as zh-hant" or zh-yue" in the next stage.

\textbf{Quality Filtering.} To achieve high-quality data, we filter documents using heuristic rules. Rather than relying on model-based evaluations like GPT-3 or LLaMA~\cite{llama, gpt3}, we implement Gopher's rule set and other heuristic criteria~\cite{rae2022gopher, 2019c4}. Thresholds in these rules, like 0.1 or 90\%, are guided by Gopher~\cite{rae2022gopher}. When strict adherence to these thresholds leads to excessive data exclusion (e.g., removing 90\% of data), we adjust them downwards. The rules include:




\begin{CJK}{UTF8}{bsmi}
(1) \textbf{Symbol-to-word ratio exceeding 0.10}: This criterion is applied to eliminate texts with an excessively high ratio of symbols to words.
(2) \textbf{Over 90\% of lines in a document commencing with a bullet point}: Documents where an overwhelming majority of lines begin with bullet points are filtered out.
(3) \textbf{Over 30\% of lines in a document terminating with ellipses}: Documents with a high frequency of lines ending in ellipses are excluded.
(4) \textbf{Word count fewer than 50 or exceeding 100,000}: Documents with extreme word counts are removed from the dataset.
(5) \textbf{Repeated n-grams, n is greater than 15}: We perform repeated 15-grams document removal. If a specific word is continuously repeated more than 15 times, we delete this document.
(6)  \textbf{Text normalization}:  For specific datasets, tailored normalization measures are implemented, such as normalizing emojis in the Openrice data and applying blacklist keywords to filter advertisements in the Apple Daily data. For example, the blacklist words are ``此回覆已被刪除'', ``撰文：阿蘭 支持蘋果深度報道，深入社區，踢爆權貴，即Like蘋果專題及調查組FB專頁！''. We also normalize the emoji into text format. For instance, 
is normalized into `thumb up`. Additionally, sequences exceeding 1000 characters are automatically truncated to meet processing requirements. Multiple line breaks and separators (e.g., \texttt{\textbackslash n} and \texttt{-}) are left into one, and during the text normalization process, blacklisted phrases, such as certain advertisements, are also included.  These procedures are critical to maintaining data consistency and cleanliness.

\end{CJK}

\begin{figure*}[t]
    \centering
    \includegraphics[width=15cm]{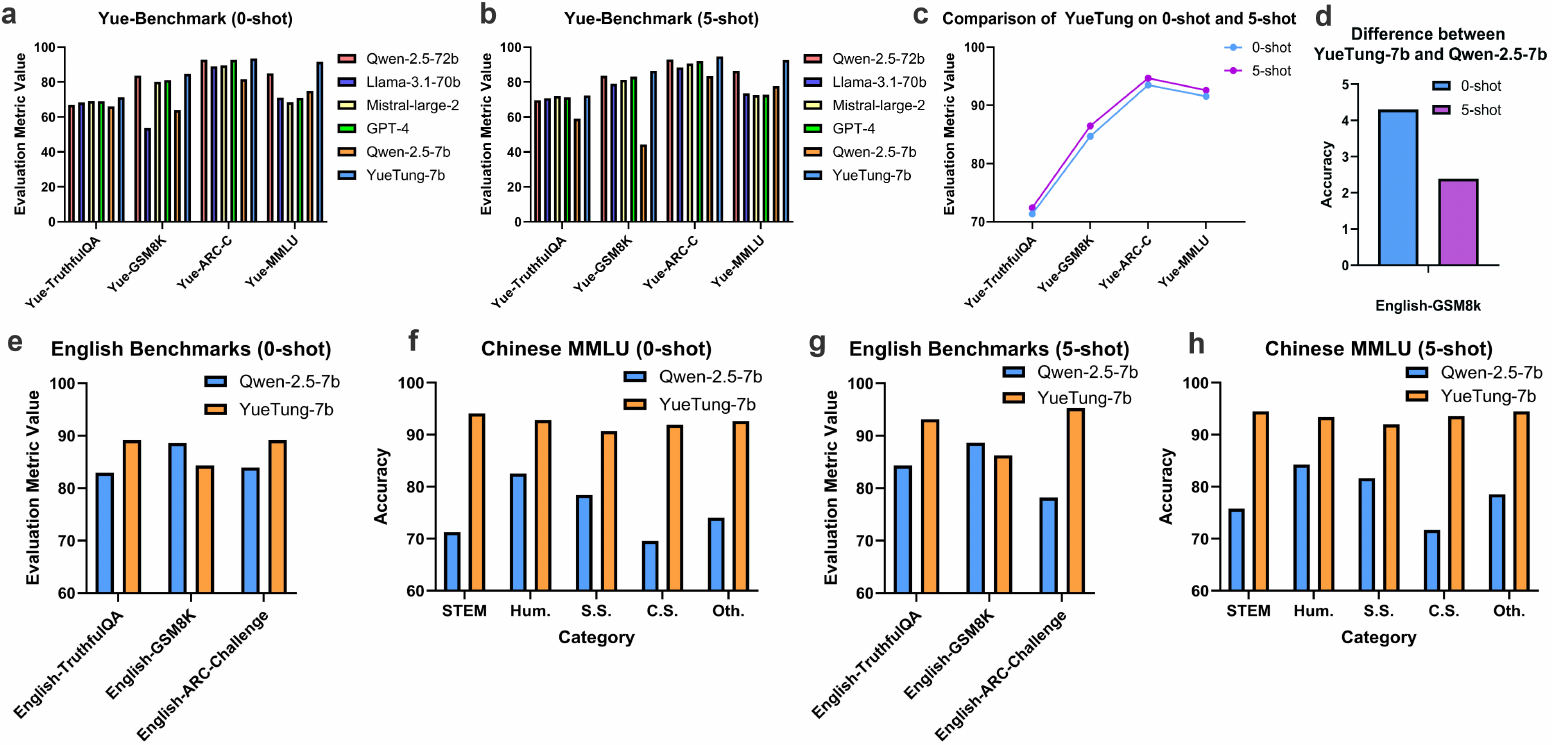}
\caption{The results of YueTung-7b and baselines on Yue-Benchmark and mainstream language benchmarks. \textbf{a} and \textbf{b} are YueTung-7b compared with representative LLMs on the Yue-Benchmark (0-shot and 5-shot). \textbf{c} is comparison of YueTung-7b on 0-shot and 5-shot. \textbf{d} is difference between YueTung-7b and Qwen-2.5-7b on the English-GSM8K. \textbf{e}, \textbf{f}, \textbf{g} and \textbf{h} are YueTung-7b compared with base model (Qwen-2.5-7b) on the mainstream language benchmarks (0-shot and 5-shot). The complete results are shown in \textbf{Table~\ref{TruthfulQA_Cant},~\ref{GSM8K_YueTung},~\ref{ARC-C_Cant},~\ref{MMLU_Cant},~\ref{GSM8K_YueTung_all},~\ref{ARC-C_Cant_all},~\ref{MMLU_Cant_all},~\ref{TruthfulQA_Eng},~\ref{GSM8K_Eng},~\ref{ARC_Eng},~\ref{CMMLU}}.}
    \label{fig:acc}
\end{figure*}


\textbf{Content filter.} Subsequently, content filter, toxicity filtering and personally identifiable information (PII) masking, are performed~\cite{dolma,Jigsaw,pii1,pii2,pii3}. The toxicity filtering utilizes two classifiers trained on the Jigsaw dataset from Dolma~\cite{dolma,Jigsaw}, specifically designed to detect and remove potentially harmful content (thresholds are set for 1e-2 and 1e-4). The filter can filter around 1\% toxic document. PII masking~\cite{pii1,pii2,pii3} is implemented using regular expressions\footnote{EMAIL\_REGEX = "[.\textbackslash\textbackslash s@,?!;:)(]*([\^{}\textbackslash\textbackslash s@]+@[\^{}\textbackslash\textbackslash s@,?!;:)\\(]+?)[.\textbackslash \textbackslash s@,?!;:)(]?[\textbackslash\textbackslash s\textbackslash n\textbackslash r]"}
\footnote{PHONE\_REGEX = "\textbackslash\textbackslash s+\textbackslash\textbackslash(?(\textbackslash\textbackslash d{3})\textbackslash\textbackslash)?[-\textbackslash\textbackslash .]*(\textbackslash\textbackslash d{3})[-. ]?(\textbackslash\textbackslash d{2,})"}
\footnote{IP\_REGEX = "(?:(?:25[0-5]|2[0-4][0-9]|[01]?[0-9][0-9]?)\textbackslash\textbackslash .){3}(?:25[0-5]|2[0-4][0-9]|[01]?[0-9][0-9]?)"}
\footnote{    URL\_REGEX = "(?i)\textbackslash b((?:https?://|www\textbackslash\textbackslash d{0,3}[.]|[a-z0-9.\textbackslash\textbackslash -]+[.][a-z]{2,4}/)(?:[\^{}\textbackslash\textbackslash s()<>]+|\textbackslash\textbackslash(([\^{}\textbackslash\textbackslash s()<>]+|(\textbackslash\textbackslash ([\^{}\textbackslash\textbackslash s()<>]+\textbackslash\textbackslash))\\)*\textbackslash\textbackslash))+(?:\textbackslash\textbackslash(([\^{}\textbackslash\textbackslash s()<>]+|(\textbackslash\textbackslash([\^{}\textbackslash\textbackslash s()<>]+\textbackslash\textbackslash)))*\textbackslash\textbackslash)|[\^{}\textbackslash\textbackslash s`!()\textbackslash\textbackslash[\textbackslash\textbackslash]{};:'\textbackslash".,\\<>?«»“”‘’]))"} to obfuscate all email addresses, phone numbers, and IP addresses into "|||IP/EMAIl/PHONE\_ADDRESS|||", thereby ensuring the protection of personal information.


\textbf{Deduplication.}
We employ MinHash and LSH techniques via the data-sketch framework to eliminate redundant documents and paragraphs~\cite{datasketch_deduplication}. Exact document deduplication removes identical documents; exact paragraph deduplication removes identical paragraphs or sentences. The deduplication threshold is typically set at 0.5 but may be adjusted to 0.6, depending on the proportion of data removed. For instance, in Common Crawl deduplication, a threshold of 0.5 removes 78.79\% of data, while 0.6 removes 44\%; thus, we choose 0.6. This time-intensive phase is essential for preserving dataset uniqueness.

This multistage data processing methodology enhances the quality of the pre-training dataset. We filter out around a 1-billion-token high-quality Cantonese corpus, with each filtering step being part of a rigorous data processing pipeline.



\begin{table}[h]\small
\centering
\begin{tabular}{l|c}
\toprule
\textbf{YueData (Pre-training)} & \textbf{Number of tokens} \\
\midrule
LIHKG & 319,604,833   \\
OpenRice & 350,050,930   \\
Apple Daily & 23,226,869 \\
HuggingFace & 402,925,178    \\
Wikipedia & 7,181,350 \\
Common Crawl & 269,777,174  \\
\bottomrule
\toprule
\textbf{YueData (SFT)} & \textbf{Number of tokens} \\
\midrule
All SFT Data & 1,289,255,036  \\
\bottomrule
\end{tabular}
\caption{YueData pre-training and supervised fine-tuning the number of tokens.}
\label{cantonesellm}
\end{table}

\subsection{Supervised Fine-Tuning Cantonese Data}

\begin{table}[h]\small
\centering
\begin{tabular}{l|c|c}
\toprule
\textbf{Models (7-8b scale)} & \textbf{Acc. (0-shot)} & \textbf{Acc. (5-shot)} \\
\midrule
Qwen-2.5-7b & 63.84 &	44.20   \\
Llama-3.1-8b & 63.91 &	61.64  \\
Yi-1.5-6b & 3.94 &	3.49  \\
Internlm-2.5-7b-chat & 65.96 &	64.67  \\
\textbf{YueTung-7b} & \textbf{84.65} & \textbf{86.46} \\
\bottomrule
\toprule
\textbf{Models (> 7-8b scale)} & \textbf{Acc. (0-shot)} & \textbf{Acc. (5-shot)} \\
\midrule
Qwen-2.5-72b & 83.62 &	83.55   \\
Mistral-large-2 & 80.14 &	81.27  \\
Llama-3.1-70b & 53.60 &	79.00  \\
Phi-3-medium & 59.29 &	63.15  \\
Gemma-2-27b & 9.70 &	2.65  \\
Yi-1.5-34b & 69.45 &	69.45  \\
Internlm-2.5-20b-chat & 71.87 &	72.33   \\
ERNIE-turbo & 14.03 	&10.92  \\
SenseChat-5 & 77.48 & 73.16 \\
Claude-3.5 & 77.79 &	81.27  \\
GLM-4 & 78.17 &	77.10  \\
ChatGPT & 23.35&	41.09 \\
GPT-4 & 81.12 &	83.02  \\
\textbf{YueTung-7b} & \textbf{84.65} & \textbf{86.46} \\
\bottomrule
\end{tabular}
\caption{Results of the comparison between answer generated by YueTung-7b and baselines in Yue-GSM8K based on 0-shot and 5-shot settings and ground truth.}
\label{GSM8K_YueTung}
\end{table}

\begin{table}[h]\small
\centering
\begin{tabular}{l|c|c}
\toprule
\textbf{Models (7-8b scale)} & \textbf{Acc. (0-shot)} & \textbf{Acc. (5-shot)} \\
\midrule
Qwen-2.5-7b & 81.64&	83.35  \\
Llama-3.1-8b & 69.00	&67.81 \\
Yi-1.5-6b & 34.59&	66.70 \\
Internlm-2.5-7b-chat & 81.21	&79.85 \\
\textbf{YueTung-7b} & \textbf{93.48} & \textbf{94.65} \\
\bottomrule
\toprule
\textbf{Models (> 7-8b scale)} & \textbf{Acc. (0-shot)} & \textbf{Acc. (5-shot)} \\
\midrule
Qwen-2.5-72b & 92.74&	92.91  \\
Mistral-large-2 & 89.50&	90.61 \\
Llama-3.1-70b & 88.98	&88.39 \\
Phi-3-medium & 77.63&	78.31 \\
Gemma-2-27b & 67.98&	55.59 \\
Yi-1.5-34b & 84.88	&86.42 \\
Internlm-2.5-20b-chat & 82.15	&82.58 \\
ERNIE-turbo & 44.41	&46.46 \\
SenseChat-5 & 88.47&	87.28 \\
Claude-3.5 & 91.55&	92.23 \\
GLM-4 & 88.90 &	88.73 \\
ChatGPT & 69.68&	70.71 \\
GPT-4 & 92.66&	92.06 \\
\textbf{YueTung-7b} & \textbf{93.48} & \textbf{94.65} \\
\bottomrule
\end{tabular}
\caption{Results of the comparison between answer generated by  YueTung-7b and baselines in Yue-ARC-C based on 0-shot and 5-shot settings and ground truth.}
\label{ARC-C_Cant}
\end{table}

\begin{table*}[h]\small
\centering
\begin{tabular}{l|c|c|c|c|c|c|c|c|c|c}
\toprule
\multirow{2}{2cm}{\textbf{Models \\ (7-8b scale)}} & \multicolumn{5}{c|}{\textbf{0-shot}} & \multicolumn{5}{c}{\textbf{5-shot}} \\
\cmidrule{2-11}
& \textbf{STEM} & \textbf{Hum.} & \textbf{S.S.} & \textbf{C.S.} & \textbf{Oth.} & \textbf{STEM} & \textbf{Hum.} & \textbf{S.S.} & \textbf{C.S.} & \textbf{Oth.}\\
\midrule
        Qwen-2.5-7b & 72.86	&81.66&	78.25&	66.56&	75.19&	78.05&	80.37&	78.99&	69.82&	78.86 \\ 
        Llama-3.1-8b & 45.96& 58.27& 56.08& 44.86& 53.70& 53.45& 58.06& 58.31& 45.86& 53.65 \\ 
        Yi-1.5-6b & 17.34& 35.98& 38.77& 32.90& 25.00& 58.53& 67.89& 66.56& 60.00& 62.05 \\ 
        Internlm-2.5-7b-chat & 64.40& 80.92& 76.80& 70.24& 75.02& 65.04& 80.84& 76.79& 70.47& 75.19 \\ 
\textbf{YueTung-7b} & \textbf{93.01} & \textbf{92.54} & \textbf{89.84} & \textbf{90.81} & \textbf{91.55} & \textbf{93.36} & \textbf{93.27} & \textbf{91.04} & \textbf{91.77} & \textbf{91.85}\\
\bottomrule
\toprule
\multirow{2}{2cm}{\textbf{Models \\ (> 7-8b scale)}} & \multicolumn{5}{c|}{\textbf{0-shot}} & \multicolumn{5}{c}{\textbf{5-shot}} \\
\cmidrule{2-11}
& \textbf{STEM} & \textbf{Hum.} & \textbf{S.S.} & \textbf{C.S.} & \textbf{Oth.} & \textbf{STEM} & \textbf{Hum.} & \textbf{S.S.} & \textbf{C.S.} & \textbf{Oth.}\\
\midrule
        Qwen-2.5-72b & 83.72&	87.88&	87.20&	80.68&	85.36&	83.89&	89.70&	88.75&	82.34&	87.42 \\ 
        Mistral-large-2 & 60.38& 76.08& 74.92& 60.19& 70.74& 68.50& 79.65& 78.84& 63.85& 71.66 \\ 
        Llama-3.1-70b & 67.32& 76.57& 76.93& 60.96& 73.56& 72.23& 78.13& 78.23& 64.16& 74.90 \\ 
        Phi-3-medium & 45.26& 61.42& 58.40& 45.65& 51.33& 49.88& 59.33& 59.35& 45.49& 53.02 \\ 
        Gemma-2-27b & 48.50& 54.05& 53.32& 36.92& 48.22& 40.62& 41.72& 43.81& 32.99& 46.03 \\ 
        Yi-1.5-34b & 68.48& 81.92& 81.74& 70.89& 79.76& 74.13& 85.12& 83.38& 78.20& 80.30 \\ 
        Internlm-2.5-20b-chat & 67.16& 81.56& 77.72& 73.05& 72.64& 66.22& 82.65& 78.42& 72.94& 74.03 \\  
        ERNIE-turbo & 43.34& 56.05& 53.97& 52.02& 44.82& 41.01& 57.66& 54.28& 49.49& 46.95 \\ 
        Sensechat-5 & 69.97& 83.21& 80.73& 73.86& 76.95& 68.98& 82.00& 79.88& 73.52& 74.77 \\ 
        Claude-3.5 & 66.47& 76.84& 78.04& 60.60& 75.98& 75.92& 81.65& 84.24& 62.83& 82.54 \\ 
GLM-4 & 64.23& 84.39& 80.06& 75.66& 75.75& 72.18& 84.20& 80.07& 76.00& 78.06 \\ 
        ChatGPT & 49.78& 58.13& 58.74& 45.46& 52.42& 60.28& 59.81& 60.61& 47.50& 54.54 \\ 
        GPT-4 & 67.68& 75.29& 77.26& 60.12& 74.46& 71.19& 76.75& 77.56& 63.50& 74.57 \\
        \textbf{YueTung-7b} & \textbf{93.01} & \textbf{92.54} & \textbf{89.84} & \textbf{90.81} & \textbf{91.55} & \textbf{93.36} & \textbf{93.27} & \textbf{91.04} & \textbf{91.77} & \textbf{91.85}\\
        
\bottomrule
\end{tabular}
\caption{Results of the comparison between texts generated by YueTung-7b and baselines in Yue-MMLU based on 0-shot and 5-shot settings and the correct texts.}
\label{MMLU_Cant}
\end{table*}


SFT data primarily originates from three sources: (1) extraction and construction from pre-trained data; (2) translation of Chinese SFT data into Cantonese; (3) collection of SFT data from GitHub.

\textbf{Extraction and Construction from Pre-trained Data:} We identify and preserve Cantonese dialogue datasets from Huggingface and Common Crawl as SFT data during pre-training. Wikipedia data relevant to knowledge retrieval is processed into a question-and-answer format (Section~\ref{prom}).

\textbf{Translating Chinese SFT Data into Cantonese:} Translating from Chinese to Cantonese is crucial for obtaining more data, as it's more rational than translating from English~\cite{jiang2024well}. We select MOSS's training data and YeungNLP's mathematical data as Chinese sources for translation. Using open-source models for large translation tasks, we choose Llama-3.1-70b based on~\cite{jiang2024well}'s comparison of LLMs in translation quality and speed. We refer to translation prompts from~\cite{jiang2024well}, conduct secondary translation, and perform partial reviews to ensure high data quality (Section~\ref{prom}). 

\textbf{Collecting Suitable SFT Data from GitHub:} We collect suitable SFT data from GitHub and incorporate it into YueTung's SFT framework.

\textbf{Data Leakage Concerns:} We focus on data leakage, ensuring pre-trained data learns Cantonese language patterns without involving test data from the Yue-Benchmark (Section~\ref{prom}).





\section{Experiment}
\subsection{Experiment Details}
Regarding model training, the YueTung model is based on Qwen-2.5-7b\footnote{\url{https://huggingface.co/Qwen/Qwen2.5-7B-Instruct}}, which is pre-trained and SFT is conducted based on the YueData dataset. Model evaluation is conducted using the four Yue-Benchmarks~\cite{jiang2024well}.

For experimental settings, we implement YueTung model with PyTorch \cite{paszke2019pytorch} on eight NVIDIA A100-80G GPUs, and train the model using AdamW optimizer \cite{loshchilov2017decoupled} with a batch size of 2. We vary the learning rate during training following \cite{vaswani2017attention}. The training time for the YueTung is about three weeks. For inference, we set the temperature as 0.2, and top-p as 1.0. 

\subsection{Evaluation and Baselines}

For Yue-TruthfulQA, we employ automatic evaluation metrics including Rouge-l~\cite{lin2004rouge}, Bleu-4~\cite{papineni2002bleu}, and BERTScore~\cite{bertscore}. For Yue-GSM8K, Yue-ARC-C, Yue-MMLU, we adopt Accuracy as evaluation metric. 


Regarding baselines, we employ LLMs from mainstream series that are either the same size as or larger than YueTung, including LLMs such as Qwen, Llama, Yi, Internlm, Mistral, Phi, Gemma, ERNIE, GLM, Sensechat, and GPT.

\section{Cantonese LLM: YueTung}

When training the YueTung-7b model, the pre-training Cantonese data contain some noisy entries, which can adversely affect the training process. To mitigate the impact of these noisy data and facilitate faster convergence, we appropriately decrease the $\beta_2$ parameter in the AdamW optimizer~\cite{loshchilov2017decoupled}. By reducing $\beta_2$, the optimizer places more emphasis on recent gradients, allowing the model to adapt more quickly and minimize the influence of noisy data. The following algorithm outlines the training procedure using the modified AdamW optimizer:

\begin{algorithm}[H]
\caption{Training YueTung model}
\begin{algorithmic}
\STATE Init params $\theta$, moments $m=0$, $v=0$, and time step $t=0$
\FOR{epochs}
    \FOR{minibatch $(X, Y)$}
        \STATE $t \leftarrow t + 1$
        \STATE Compute grad $g = \nabla_\theta L(\theta; X, Y)$
        \STATE $m \leftarrow \beta_1 m + (1 - \beta_1) g$
        \STATE $v \leftarrow \beta_2 v + (1 - \beta_2) g^2$
        \STATE $\hat{m} \leftarrow m / (1 - \beta_1^t)$
        \STATE $\hat{v} \leftarrow v / (1 - \beta_2^t)$
        \STATE Update $\theta \leftarrow \theta - \alpha \left( \frac{\hat{m}}{\sqrt{\hat{v}} + \epsilon} + \lambda \theta \right)$
    \ENDFOR
\ENDFOR
\end{algorithmic}
\end{algorithm}

where $L(\theta; X, Y)$ is the loss function (during pre-training, $Y$ may be omitted); decreasing $\beta_2$ allows the optimizer to adapt more quickly to recent gradients, mitigating the impact of noisy data; initial moments $m$ and $v$ are zero with bias correction applied; hyper-parameters include $\alpha$, $\beta_1$, $\beta_2$, $\lambda$, and $\epsilon$; weight decay $\lambda \theta$ is included directly in the update; the algorithm applies to both pre-training and supervised fine-tuning stages.

Since LoRA can significantly reduce computational and memory requirements~\cite{wang2024mos, wang2024prolora, hu2022lora}, we incorporate LoRA to enhance training efficiency.

\section{Results and Analysis}
Section~\ref{analysis} for a more detailed analysis.

\subsection{Cantonese Benchmarks}

In Yue-TruthfulQA (Table~\ref{TruthfulQA_Cant}), YueTung-7b achieves Rouge-l scores of 33.95\% (zero-shot) and 35.12\% (five-shot), outperforming all baseline models, including GPT-4 and ChatGPT. Its highest BERTScore indicates superior semantic similarity to the ground truth. For Yue-GSM8K (Table~\ref{GSM8K_YueTung}), YueTung-7b attains accuracies of 84.65\% (zero-shot) and 86.46\% (five-shot), significantly exceeding other 7B to 8B models and even surpassing larger models like GPT-4, highlighting strong reasoning capabilities in Cantonese problem-solving. On Yue-ARC-C (Table~\ref{ARC-C_Cant}), YueTung-7b achieves accuracies of 93.48\% (zero-shot) and 94.65\% (five-shot), outperforming all other models, including GPT-4 and GPT-4o, indicating proficiency in challenging Cantonese comprehension tasks. In Yue-MMLU (Table~\ref{MMLU_Cant}), YueTung-7b consistently achieves accuracies above 89\%, peaking at 93.36\% in STEM (five-shot), leading over larger LLMs like Qwen-2.5-72b and GPT-4, underscoring its comprehensive knowledge base in Cantonese.

YueTung-7b achieves SOTA performance across all Cantonese benchmarks. Its superior results demonstrate strong Cantonese language proficiency, excelling in context understanding, reasoning, and knowledge retrieval. The significant performance gap suggests that the high-quality Cantonese dataset (YueData) and tailored training strategies contribute greatly to its success. YueTung-7b's ability to outperform larger models like GPT-4 emphasizes the importance of language-specific data in low-resource languages.

\subsection{Mainstream Language Benchmarks}

In English-TruthfulQA (Table~\ref{TruthfulQA_Eng}), YueTung-7b achieves Rouge-l scores of 37.41\% (zero-shot) and 63.50\% (five-shot), competitive with larger models like ChatGPT and GPT-4. Its high BERTScore indicates effective cross-lingual knowledge transfer. On English-GSM8K (Table~\ref{GSM8K_Eng}), YueTung-7b attains accuracies of 84.32\% (zero-shot) and 86.26\% (five-shot), indicating robust mathematical reasoning in English. For English-ARC Challenge (Table~\ref{ARC_Eng}), YueTung-7b achieves accuracies of 89.15\% (zero-shot) and 95.25\% (five-shot), surpassing several larger models, demonstrating effective handling of English multiple-choice questions. On CMMLU in Standard Chinese (Table~\ref{CMMLU}), YueTung-7b achieves overall accuracies of 92.63\% (zero-shot) and 94.49\% (five-shot), outperforming all other models, including large-scale ones like Qwen-2.5-72b and GPT-4, indicating enhanced capabilities in closely related languages.

YueTung-7b excels in Cantonese and demonstrates strong cross-lingual abilities in English and Mandarin. This suggests Cantonese data fosters robust language understanding beyond Cantonese, highlighting the potential of leveraging Cantonese data to enhance overall LLM performance.

\section{Conclusion}
This paper addresses the challenges of Cantonese as a low-resource language by constructing a high-quality corpus exceeding 2 billion tokens, collected from diverse sources and rigorously processed for training LLM (YueTung). YueTung is refined through supervised fine-tuning on curated Cantonese tasks, achieves SOTA performance on four benchmarks and shows improved results on other mainstream language tasks. 





\section*{Limitations}

While YueTung-7b exhibits exceptional performance, there are limitations to our current work. 
For instance, the YueData corpus, though extensive, predominantly comprises text from Hong Kong-specific sources. 
As a result, the model may be biased toward the linguistic styles, idioms, and colloquialisms prevalent in Hong Kong Cantonese, potentially limiting its generalizability to other Cantonese dialects spoken in different regions.

In addition, despite our rigorous data processing efforts, including language filtering, quality filtering, content filtering, and deduplication, some noise and biases may persist in the dataset. The complexities of Cantonese, such as code-switching with English and the use of non-standard characters, pose challenges that may affect the model's performance in certain contexts or with highly informal language.



\section*{Ethics Statement}

This paper does not involve ethics-related issues.

\bibliography{custom}

\appendix

\section{Appendix}
\label{sec:appendix}

\subsection{Results Analysis}
\label{analysis}
\subsubsection{Cantonese Benchmarks}

We evaluate YueTung-7b on four Cantonese benchmarks: Yue-TruthfulQA, Yue-GSM8K, Yue-ARC-C, and Yue-MMLU. The results are summarized in Tables~\ref{TruthfulQA_Cant},~\ref{GSM8K_YueTung},~\ref{ARC-C_Cant}, and~\ref{MMLU_Cant}, respectively.

About Yue-TruthfulQA, As shown in Table~\ref{TruthfulQA_Cant}, YueTung-7b achieves a Rouge-l score of 33.95\% in the zero-shot setting and 35.12\% in the five-shot setting, outperforming all baseline models of similar and larger scales. 
Notably, YueTung-7b substantially surpasses GPT-4 and ChatGPT, which achieve Rouge-l scores of 19.47\% and 25.07\% in the zero-shot setting, respectively. 
The BERTScore of YueTung-7b is also the highest among all models, indicating superior semantic similarity to the ground truth. 
These results demonstrate YueTung-7b's ability to generate truthful and coherent responses in Cantonese.

About Yue-GSM8K, Table~\ref{GSM8K_YueTung} presents the accuracy results on Yue-GSM8K, a mathematical reasoning benchmark. 
YueTung-7b attains an accuracy of 84.65\% in the zero-shot setting and 86.46\% in the five-shot setting. 
This significantly exceeds the performance of other 7B to 8B scale models, such as Qwen-2.5-7b, which achieves 63.84\% and 44.20\% accuracy, respectively. 
YueTung-7b also outperforms larger models like GPT-4 and GPT-4o, highlighting its strong reasoning capabilities in Cantonese mathematical problem-solving.

About Yue-ARC-C, on the Yue-ARC-C benchmark, which tests knowledge and reasoning in multiple-choice questions, YueTung-7b achieves accuracies of 93.48\% (zero-shot) and 94.65\% (five-shot), as shown in Table~\ref{ARC-C_Cant}. This places it ahead of all other models, including GPT-4o and GPT-4, which achieve accuracies around 92\%. The substantial margin indicates YueTung-7b's proficiency in handling challenging Cantonese comprehension tasks.

About Yue-MMLU, YueTung-7b's performance on Yue-MMLU is detailed in Table~\ref{MMLU_Cant}. 
Across all categories—STEM, Humanities, Social Sciences, Computer Science, and Others—YueTung-7b consistently achieves accuracies above 89\%, with the highest being 93.36\% in the STEM category for the five-shot setting. 
Compared to other models, YueTung-7b exhibits a remarkable lead, outperforming larger models like Qwen-2.5-72b and GPT-4 by a significant margin. 
This consistent performance across diverse subjects underscores YueTung-7b's comprehensive knowledge base and understanding of Cantonese.

\paragraph{Analysis}
The experimental results on Cantonese benchmarks demonstrate that YueTung-7b achieves SOTA performance across all evaluated tasks. Its superior results in both zero-shot and five-shot settings indicate that the model not only has a strong grasp of the Cantonese language but also excels in understanding context, reasoning, and knowledge retrieval. The substantial performance gap between YueTung-7b and other models of similar scale suggests that the high-quality Cantonese dataset (YueData) and the tailored training strategies significantly contribute to its success.

Moreover, YueTung-7b's ability to outperform much larger models like GPT-4 emphasizes the importance of language-specific data in low-resource languages. The results validate our approach of focusing on data quality and appropriate training techniques to enhance model performance in Cantonese NLP tasks.

\subsubsection{Mainstream Language Benchmarks}

To assess the generalization capabilities of YueTung-7b beyond Cantonese, we evaluate the model on mainstream language benchmarks, including English and Standard Chinese tasks. The results are presented in Tables~\ref{TruthfulQA_Eng},~\ref{GSM8K_Eng},~\ref{ARC_Eng},~\ref{CMMLU}.

About English-TruthfulQA, in Table~\ref{TruthfulQA_Eng}, YueTung-7b achieves a Rouge-l score of 37.41\% in the zero-shot setting and an impressive 63.50\% in the five-shot setting on the English-TruthfulQA benchmark. These scores are competitive with larger models like ChatGPT, which scores 37.81\% (zero-shot) and 50.43\% (five-shot), and GPT-4, which achieves 19.58\% (zero-shot) and 53.18\% (five-shot). YueTung-7b's high BERTScore indicates strong semantic similarity to the ground truth, suggesting effective cross-lingual transfer of knowledge.

About English-GSM8K, Table~\ref{GSM8K_Eng} shows that YueTung-7b attains accuracies of 84.32\% (zero-shot) and 86.26\% (five-shot) on the English-GSM8K benchmark. While it slightly lags behind top-performing models like Qwen-2.5-72b and GPT-4o, which achieve accuracies above 93\%, YueTung-7b's performance is notable given its smaller parameter size and focus on Cantonese data. The results indicate that YueTung-7b retains robust mathematical reasoning abilities in English.

About English-ARC Challenge, on the English-ARC Challenge benchmark (Table~\ref{ARC_Eng}), YueTung-7b achieves accuracies of 89.15\% (zero-shot) and 95.25\% (five-shot). This performance is competitive with larger models and surpasses several, such as Qwen-2-72b and Llama-3-70b. YueTung-7b's strong results suggest that it can effectively handle English multiple-choice questions, demonstrating cross-lingual generalizability.

About CMMLU, Table~\ref{CMMLU} presents YueTung-7b's performance on the CMMLU benchmark in Standard Chinese. The model achieves high accuracies across all categories, with overall accuracies of 92.63\% (zero-shot) and 94.49\% (five-shot). YueTung-7b outperforms all other models, including large-scale models like Qwen-2.5-72b and GPT-4. This indicates that training on comprehensive Cantonese data enhances the model's capabilities in closely related languages like Standard Chinese.

\paragraph{Analysis}
YueTung-7b's performance on mainstream language benchmarks reveals that the model not only excels in Cantonese but also demonstrates strong cross-lingual abilities in English and Standard Chinese. The model consistently performs well across different tasks and settings, suggesting that the high-quality Cantonese data contributes to a robust underlying language understanding that generalizes beyond Cantonese.

These findings highlight the potential of leveraging low-resource language data to improve overall model performance. YueTung-7b's ability to compete with or surpass larger models on mainstream benchmarks underscores the effectiveness of our data collection and training approach.

\subsection{Evaluation Tools}

\begin{itemize}
    \item Rouge-l: from rouge\_metric import PyRouge
    \item Bleu-4: from nltk.translate.bleu\_score import sentence\_bleu, SmoothingFunction
    \item BERTScore: bert-base-multilingual-cased \& roberta-large
\end{itemize}

\subsection{SFT Data Construction}
\label{prom}
\subsubsection{Cons. Prompts}
We can directly extract readily available dialogue data from Huggingface and Common Crawl to construct it into the SFT data format.

Wikipedia data can be structured into the format of SFT dialogues based on the following prompt.

\begin{CJK}{UTF8}{bsmi}
Human: 請問<concept>係咩？ \textbf{\textbackslash n} Assistant: <content>。
\end{CJK}

\textbf{Translation:}

Human: What is <concept>? \textbf{\textbackslash n} Assistant: <content>.

\subsubsection{SFT Data Translation Prompts}

Our first round of prompt from Chinese to Cantonese:

\begin{CJK}{UTF8}{bsmi}
你係一位專業嘅中文翻譯粵語嘅翻譯員。你嘅任務係準確噉將所提供嘅中文文本翻譯成粵語，同時保留原文嘅意思、語氣同格式。嚴格遵守以下規則：\textbf{\textbackslash n} 1. 只輸出翻譯結果，唔加入任何解釋、步驟或額外內容。\textbf{\textbackslash n} 2. 保持原有嘅段落結構同標點。\textbf{\textbackslash n} 3. 唔重複、遺漏或改動原文嘅任何部分。\textbf{\textbackslash n} 4. 保持數字或公式不變，唔進行任何計算或修改。\textbf{\textbackslash n}\textbf{\textbackslash n} 中文文本：\textbf{\textbackslash n}
\end{CJK}

\textbf{Translation:}

You are a professional translator specializing in translating from Chinese into Cantonese. Your task is to accurately translate the provided Chinese text into Cantonese while preserving the original meaning, tone, and format. Strictly adhere to the following rules: \textbf{\textbackslash n} 1. Only output the translation result, without adding any explanations, steps, or additional content. \textbf{\textbackslash n} 2. Maintain the original paragraph structure and punctuation. \textbf{\textbackslash n} 3. Do not repeat, omit, or alter any part of the original text. \textbf{\textbackslash n} 4. Keep numbers or formulas unchanged, without performing any calculations or modifications. \textbf{\textbackslash n}\textbf{\textbackslash n} Chinese text: \textbf{\textbackslash n}

The second round of prompt from Chinese to Cantonese:

\begin{CJK}{UTF8}{bsmi}
\textbf{(System prompt)} You are a professional translator specialized in translating Chinese into Cantonese. Your task is to refine and provide a more accurate Cantonese translation based on the original Chinese text and the previous translation result. Please strictly follow these guidelines: \textbf{\textbackslash n}\textbf{\textbackslash n} 1. Only output the corrected Cantonese translation. Do NOT adding any explanations, steps, calculations, inferences, or extra content. \textbf{\textbackslash n} 2. Preserve the original paragraph structure and punctuation. \textbf{\textbackslash n} 3. Do not repeat, omit, or alter any part of the original text. \textbf{\textbackslash n} 4. Keep numbers and formulas unchanged, without performing any calculations or modifications. \textbf{\textbackslash n}\textbf{\textbackslash n}

\textbf{(Human few shot 1)} Example 1:\textbf{\textbackslash n} Original Chinese Text: 题目：'小明每天早上花费10分钟时间走到学校，如果小明家距离学校2公里，那么他每分钟走多少米？' \textbf{\textbackslash n} Cantonese Translation: '題目：小明每日朝早花10分鐘時間行去學校，如果小明屋企距離學校2公里，咁佢每分鐘行幾多米？' \textbf{\textbackslash n} Example 2: \textbf{\textbackslash n} Original Chinese Text: '题目：今天小明骑自行车从家到学校用了20分钟，回家用了25分钟。如果小明在上学和回家的路上的速度一样，那么他从家到学校的距离是学校到家的距离的百分之几？' \textbf{\textbackslash n} Cantonese Translation: '題目：今日小明踩單車由屋企去學校用咗20分鐘，返屋企用咗25分鐘。如果小明喺返學同返屋企嘅路上嘅速度係一樣嘅，咁佢由屋企去學校嘅距離係學校返屋企嘅距離嘅百分之幾？' \textbf{\textbackslash n} Example 3: \textbf{\textbackslash n} Original Chinese Text: '题目：\textbf{\textbackslash n} 鹿妈妈买了24个苹果，她想平均分给她的3只小鹿吃，每只小鹿可以分到几个苹果？' \textbf{\textbackslash n} Cantonese Translation: '題目：\textbf{\textbackslash n} 鹿媽媽買咗24個蘋果，佢想平均分俾佢嘅3隻小鹿食，每隻小鹿可以分到幾多個蘋果？' \textbf{\textbackslash n}

\textbf{(Human few shot 2)} Example 1: \textbf{\textbackslash n} Original Chinese Text: '这是一个关于速度、路程、时间的数学问题。我们可以通过公式：速度＝路程÷时间 来解决。 \textbf{\textbackslash n} 因为小明每天早上走2公里，所以他的路程为2千米。而他每天早上要花费10分钟时间走到学校，因此他的时间为10分钟，即600秒。 \textbf{\textbackslash n} 所以小明每分钟走的距离为 2公里 / 600秒 = 0.0033公里/秒 或 3.3米/秒。 \textbf{\textbackslash n} 答案：小明每分钟走3.3米。' \textbf{\textbackslash n} Cantonese Translation: '呢個係一條關於速度、路程、時間嘅數學題目。我哋可以通過公式：速度＝路程÷時間 來解決。 \textbf{\textbackslash n} 因為小明每日朝早行2公里，所以佢嘅路程為2公里。而佢每日朝早要花10分鐘時間行去學校，因此佢嘅時間為10分鐘，即係600秒。 \textbf{\textbackslash n} 所以小明每分鐘行嘅距離係 2公里 / 600秒 = 0.0033公里/秒 或 3.3米/秒。 \textbf{\textbackslash n} 答案：小明每分鐘行3.3米。' \textbf{\textbackslash n} Example 2: \textbf{\textbackslash n} Original Chinese Text: '假设小明家到学校的距离为x千米，根据速度等于路程除以时间的公式，可以得出小明的速度为：家到学校的速度 = x / 20，学校到家的速度 = x / 25。因为小明在上学和回家的路上的速度一样，所以有：x / 20 = x / 25，解出 x = 5/4 千米。 \textbf{\textbackslash n} 因此，家到学校的距离是学校到家的距离的百分之几，可以通过求比值得到：x / (5/4)x = 4/5 = 0.8，即小明从家到学校的距离是学校到家的距离的百分之80。' \textbf{\textbackslash n} Cantonese Translation: '假設小明屋企去學校嘅距離係x公里，根據速度等於路程除以時間嘅公式，可以得出小明嘅速度為：屋企去學校嘅速度 = x / 20，學校返屋企嘅速度 = x / 25。因為小明喺返學同返屋企嘅路上嘅速度係一樣，所以有：x / 20 = x / 25，解出 x = 5/4 公里。 \textbf{\textbackslash n} 因此，屋企去學校嘅距離係學校返屋企嘅距離嘅百分之幾，可以通過求比值得出：x / (5/4)x = 4/5 = 0.8，即小明由屋企去學校嘅距離係學校返屋企嘅距離嘅百分之80。' \textbf{\textbackslash n} Example 3: \textbf{\textbackslash n} Original Chinese Text: '鹿妈妈买了24个苹果，平均分给3只小鹿吃，那么每只小鹿可以分到的苹果数就是总苹果数除以小鹿的只数。 \textbf{\textbackslash n} 24÷3=8 \textbf{\textbackslash n} 每只小鹿可以分到8个苹果。所以，答案是每只小鹿可以分到8个苹果。' \textbf{\textbackslash n} Cantonese Translation: '鹿媽媽買咗24個蘋果，平均分俾3隻小鹿食，咁每隻小鹿可以分到嘅蘋果數就係總蘋果數除以小鹿嘅隻數。 \textbf{\textbackslash n} 24÷3=8 \textbf{\textbackslash n} 每隻小鹿可以分到8個蘋果。所以，答案係每隻小鹿可以分到8個蘋果。' \textbf{\textbackslash n}
\end{CJK}

\textbf{Translation:}

\textbf{(Human few shot 1)} Example 1: \textbf{\textbackslash n} Original Chinese Text: 'Question: "Xiao Ming spends 10 minutes every morning walking to school. If his home is 2 kilometers away from school, how many meters does he walk per minute?"' \textbf{\textbackslash n} Cantonese Translation: 'Question: Xiao Ming spends 10 minutes every morning walking to school. If his home is 2 kilometers away from the school, how many meters does he walk per minute?' \textbf{\textbackslash n} Example 2: \textbf{\textbackslash n} Original Chinese Text: 'Question: "Today Xiao Ming rode his bicycle from home to school in 20 minutes and it took him 25 minutes to return. If his speed was the same going to and coming from school, what percentage of the distance from home to school is the distance from school to home?"' \textbf{\textbackslash n} Cantonese Translation: 'Question: Today Xiao Ming cycled from home to school in 20 minutes, and it took him 25 minutes to return. If his speed on both the trip to school and the return home was the same, then what percentage of the distance from home to school is the distance from school to home?' \textbf{\textbackslash n} Example 3: \textbf{\textbackslash n} Original Chinese Text: 'Question: \textbf{\textbackslash n} "Deer Mother bought 24 apples, she wants to evenly distribute them to her 3 fawns, how many apples does each fawn get?"' \textbf{\textbackslash n} Cantonese Translation: 'Question: \textbf{\textbackslash n} Deer Mother bought 24 apples, she wants to distribute them evenly among her 3 fawns, how many apples does each fawn get?' \textbf{\textbackslash n} 

\textbf{(Human few shot 2)} Example 1: \textbf{\textbackslash n} Original Chinese Text: 'This is a math question about speed, distance, and time. We can solve it using the formula: Speed = Distance ÷ Time. \textbf{\textbackslash n} Since Xiao Ming walks 2 kilometers every morning, his distance is 2 kilometers. Since he spends 10 minutes walking to school each morning, his time is 10 minutes, which is 600 seconds. \textbf{\textbackslash n} Therefore, the distance Xiao Ming walks per minute is 2 kilometers / 600 seconds = 0.0033 kilometers/second or 3.3 meters/second. \textbf{\textbackslash n} Answer: Xiao Ming walks 3.3 meters per minute.' \textbf{\textbackslash n} Cantonese Translation: 'This is a math problem about speed, distance, and time. We can solve it using the formula: Speed = Distance ÷ Time. \textbf{\textbackslash n} Since Xiao Ming walks 2 kilometers every morning, his distance is 2 kilometers. Since he spends 10 minutes walking to school each morning, his time is 10 minutes, or 600 seconds. \textbf{\textbackslash n} Thus, the distance Xiao Ming walks per minute is 2 kilometers / 600 seconds = 0.0033 kilometers/second or 3.3 meters/second. \textbf{\textbackslash n} Answer: Xiao Ming walks 3.3 meters per minute.' \textbf{\textbackslash n} Example 2: \textbf{\textbackslash n} Original Chinese Text: 'Assuming the distance from Xiao Ming's home to school is x kilometers, based on the formula that speed equals distance divided by time, Xiao Ming's speed can be calculated as: speed from home to school = x / 20, speed from school to home = x / 25. Since Xiao Ming's speed to and from school is the same, we have: x / 20 = x / 25, solving x = 5/4 kilometers. \textbf{\textbackslash n} Therefore, the percentage of the distance from home to school that is the distance from school to home can be found by calculating the ratio: x / (5/4)x = 4/5 = 0.80, meaning the distance from home to school is 80\% of the distance from school to home.' \textbf{\textbackslash n} Cantonese Translation: 'Assuming the distance from Xiao Ming's home to school is x kilometers, based on the formula that speed equals distance divided by time, we can calculate Xiao Ming's speed as: speed from home to school = x / 20, speed from school to home = x / 25. Since Xiao Ming's speed to and from school is the same, we have: x / 20 = x / 25, solving x = 5/4 kilometers. \textbf{\textbackslash n} Therefore, the percentage of the distance from home to school that is the distance from school to home can be calculated by determining the ratio: x / (5/4)x = 4/5 = 0.80, meaning the distance from home to school is 80\% of the distance from school to home.' \textbf{\textbackslash n} Example 3: \textbf{\textbackslash n} Original Chinese Text: 'Deer Mother bought 24 apples and wants to divide them equally among her 3 fawns, so the number of apples each fawn gets is the total number of apples divided by the number of fawns. \textbf{\textbackslash n} 24÷3=8 \textbf{\textbackslash n} Each fawn gets 8 apples. Therefore, the answer is each fawn gets 8 apples.' \textbf{\textbackslash n} Cantonese Translation: 'Deer Mother bought 24 apples and wants to divide them equally among her 3 fawns, so the number of apples each fawn gets is the total number of apples divided by the number of fawns. \textbf{\textbackslash n} 24÷3=8 \textbf{\textbackslash n} Each fawn gets 8 apples. Therefore, the answer is each fawn gets 8 apples.' \textbf{\textbackslash n}

\subsubsection{Data Leakage Concerns}
We also focus on data leakage, given that pre-trained data primarily learns Cantonese language patterns and does not involve test data from the Yue-Benchmark.

Our concern is whether there is data in the SFT that resembles that in the Yue-Benchmark. Due to computational speed limitations, we used the Bleu metric to identify data most similar to the Yue-Benchmark test data and performed Bleu and BERTScore linguistic similarity calculations, ultimately observing a Bleu score of 0.08 and a BERTScore of 0.32.

\subsection{Conclusion}

This paper addresses the challenges of Cantonese as a low-resource language by constructing a high-quality corpus exceeding 2 billion tokens, collected from diverse sources and rigorously processed for training LLM (YueTung). YueTung is refined through supervised fine-tuning on curated Cantonese tasks, achieves SOTA performance on four benchmarks and shows improved results on other mainstream language tasks.

\subsection{All Results}

\subsubsection{Cantonese Benchmarks}

\begin{table}[h]\small
\centering
\begin{tabular}{l|c|c}
\toprule
\textbf{Models (7-8b scale)} & \textbf{Acc. (0-shot)} & \textbf{Acc. (5-shot)} \\
\midrule
Qwen-7b & 0.68    & 6.75  \\
Qwen-1.5-7b & 36.62 	&26.31   \\
Qwen-2-7b & 50.49 &	61.11  \\
Qwen-2.5-7b & 63.84 &	44.20   \\
Llama-2-7b & 0.83 	&1.82  \\
Llama-3-8b & 52.46 &	49.66  \\
Llama-3.1-8b & 63.91 &	61.64  \\
Yi-6b & 2.12 &	10.16  \\
Yi-1.5-6b & 3.94 &	3.49  \\
Internlm-7b & 4.55 &	9.48  \\
Internlm-2-7b-chat & 56.41 &	48.67  \\
Internlm-2-7b & 11.37 	&23.96  \\
Internlm-2.5-7b-chat & 65.96 &	64.67  \\
Internlm-2.5-7b & 56.79 &	42.99  \\
\textbf{YueTung-7b} & \textbf{84.65} & \textbf{86.46} \\
\bottomrule
\toprule
\textbf{Models (> 7-8b scale)} & \textbf{Acc. (0-shot)} & \textbf{Acc. (5-shot)} \\
\midrule
Qwen-1.5-110b & 54.89 &	58.30  \\
Qwen-2-72b & 77.86 &	77.71  \\
Qwen-2.5-72b & 83.62 &	83.55   \\
Mistral-8x22b & 65.20 &	66.19  \\
Mistral-large-2 & 80.14 &	81.27  \\
Llama-3-70b & 73.62 &	75.66  \\
Llama-3.1-70b & 53.60 &	79.00  \\
Phi-3-medium & 59.29 &	63.15  \\
Gemma-2-27b & 9.70 &	2.65  \\
Yi-1.5-34b & 69.45 &	69.45  \\
Internlm-2-20b & 12.81 &	8.87  \\
Internlm-2-20b-chat & 60.42 &	59.21  \\
Internlm-2.5-20b-chat & 71.87 &	72.33   \\
Internlm-2.5-20b & 45.03 &	61.41  \\
ERNIE-turbo & 14.03 	&10.92  \\
ERNIE-Speed & 28.81 	&28.28  \\
ERNIE-Lite & 54.81 	&32.15  \\
ERNIE-Tiny & 2.73 &	3.94  \\
SenseChat-5 & 77.48 & 73.16 \\
Claude-3.5 & 77.79 &	81.27  \\
GLM-4 & 78.17 &	77.10  \\
ChatGPT & 23.35&	41.09 \\
GPT-4o & 83.24 &	83.40  \\
GPT-4 & 81.12 &	83.02  \\
\textbf{YueTung-7b} & \textbf{84.65} & \textbf{86.46} \\
\bottomrule
\end{tabular}
\caption{\textbf{All results} of the comparison between answer generated by YueTung-7b and baselines in Yue-GSM8K based on 0-shot and 5-shot settings and ground truth.}
\label{GSM8K_YueTung_all}
\end{table}

\begin{table}[h]\small
\centering
\begin{tabular}{l|c|c}
\toprule
\textbf{Models (7-8b scale)} & \textbf{Acc. (0-shot)} & \textbf{Acc. (5-shot)} \\
\midrule
Qwen-7b & 11.02	&14.60  \\
Qwen-1.5-7b & 65.24	&67.55  \\
Qwen-2-7b & 79.08&	78.39  \\
Qwen-2.5-7b & 81.64&	83.35  \\
Llama-2-7b & 23.57	&34.24  \\
Llama-3-8b & 70.11&	53.80 \\
Llama-3.1-8b & 69.00	&67.81 \\
Yi-6b & 31.00&	66.01 \\
Yi-1.5-6b & 34.59&	66.70 \\
Internlm-7b & 44.75&	55.34 \\
Internlm-2-7b & 44.75	&55.34 \\
Internlm-2.5-7b-chat & 81.21	&79.85 \\
Internlm-2.5-7b & 77.37&	77.37 \\
\textbf{YueTung-7b} & \textbf{93.48} & \textbf{94.65} \\
\bottomrule
\toprule
\textbf{Models (> 7-8b scale)} & \textbf{Acc. (0-shot)} & \textbf{Acc. (5-shot)} \\
\midrule
Qwen-1.5-110b & 88.64	&90.09 \\
Qwen-2-72b & 88.64	&88.56  \\
Qwen-2.5-72b & 92.74&	92.91  \\
Mistral-8x22b & 76.09&	76.09 \\
Mistral-large-2 & 89.50&	90.61 \\
Llama-3-70b & 85.06&	84.97 \\
Llama-3.1-70b & 88.98	&88.39 \\
Phi-3-medium & 77.63&	78.31 \\
Gemma-2-27b & 67.98&	55.59 \\
Yi-1.5-34b & 84.88	&86.42 \\
Internlm-2.5-20b-chat & 82.15	&82.58 \\
Internlm-2.5-20b & 84.29&	76.94 \\
ERNIE-turbo & 44.41	&46.46 \\
ERNIE-Speed & 74.47	&74.04 \\
ERNIE-Lite & 72.25	&77.28 \\
ERNIE-Tiny & 34.67&	32.88 \\
SenseChat-5 & 88.47&	87.28 \\
Claude-3.5 & 91.55&	92.23 \\
GLM-4 & 88.90 &	88.73 \\
ChatGPT & 69.68&	70.71 \\
GPT-4o & 91.97&	94.45 \\
GPT-4 & 92.66&	92.06 \\
\textbf{YueTung-7b} & \textbf{93.48} & \textbf{94.65} \\
\bottomrule
\end{tabular}
\caption{\textbf{All results} of the comparison between answer generated by  YueTung-7b and baselines in Yue-ARC-C based on 0-shot and 5-shot settings and ground truth.}
\label{ARC-C_Cant_all}
\end{table}

\begin{table*}\small
\centering
 \begin{tabular}{l|c|c|c|c|c|c|c|c|c|c}
\toprule
\multirow{2}{2cm}{\textbf{Models \\ (7-8b scale)}} & \multicolumn{5}{c|}{\textbf{0-shot}} & \multicolumn{5}{c}{\textbf{5-shot}} \\
\cmidrule{2-11}
& \textbf{STEM} & \textbf{Hum.} & \textbf{S.S.} & \textbf{C.S.} & \textbf{Oth.} & \textbf{STEM} & \textbf{Hum.} & \textbf{S.S.} & \textbf{C.S.} & \textbf{Oth.}\\
\midrule
Qwen-7b & 10.10 & 12.95& 12.12& 11.61& 7.96& 9.98& 15.96& 14.48& 13.33& 13.26 \\ 
        Qwen-1.5-7b & 46.28& 61.65& 56.57& 50.02& 53.00& 60.14& 70.09& 65.55& 58.31& 65.02 \\ 
        Qwen-2-7b & 70.06& 81.04& 80.07& 69.54& 76.04& 74.08& 80.45& 80.70& 73.70& 79.52 \\ 
        Qwen-2.5-7b & 72.86	&81.66&	78.25&	66.56&	75.19&	78.05&	80.37&	78.99&	69.82&	78.86 \\ 
        Llama-2-7b & 23.34& 23.84& 23.76& 22.78& 24.52& 27.48& 30.40& 31.76& 28.90& 24.38 \\ 
        Llama-3-8b & 49.13& 59.30& 56.51& 47.53& 53.72& 44.04& 58.47& 53.94& 46.24& 52.55 \\ 
        Llama-3.1-8b & 45.96& 58.27& 56.08& 44.86& 53.70& 53.45& 58.06& 58.31& 45.86& 53.65 \\ 
        Yi-6b & 36.46& 67.62& 57.32& 57.42& 50.06& 58.11& 72.14& 68.40& 60.56& 68.46 \\ 
        Yi-1.5-6b & 17.34& 35.98& 38.77& 32.90& 25.00& 58.53& 67.89& 66.56& 60.00& 62.05 \\ 
        Internlm-7b & 31.90 & 48.79& 44.03& 41.14& 39.82& 39.84& 51.74& 50.06& 43.60& 42.32 \\ 
        Internlm-2-7b & 51.69& 70.92& 64.71& 59.31& 58.93& 53.11& 68.51& 62.68& 59.77& 58.14 \\ 
        Internlm-2.5-7b-chat & 64.40& 80.92& 76.80& 70.24& 75.02& 65.04& 80.84& 76.79& 70.47& 75.19 \\ 
        Internlm-2.5-7b & 65.34& 82.43& 79.24& 73.11& 74.15& 66.73& 81.06& 77.80& 71.65& 75.37 \\ 
\textbf{YueTung-7b} & \textbf{93.01} & \textbf{92.54} & \textbf{89.84} & \textbf{90.81} & \textbf{91.55} & \textbf{93.36} & \textbf{93.27} & \textbf{91.04} & \textbf{91.77} & \textbf{91.85}\\
\bottomrule
\toprule
\multirow{2}{2cm}{\textbf{Models \\ (> 7-8b scale)}} & \multicolumn{5}{c|}{\textbf{0-shot}} & \multicolumn{5}{c}{\textbf{5-shot}} \\
\cmidrule{2-11}
& \textbf{STEM} & \textbf{Hum.} & \textbf{S.S.} & \textbf{C.S.} & \textbf{Oth.} & \textbf{STEM} & \textbf{Hum.} & \textbf{S.S.} & \textbf{C.S.} & \textbf{Oth.}\\
\midrule
        Qwen-1.5-110b & 75.07& 88.48& 83.89& 80.57& 82.14& 79.96& 88.12& 88.75& 84.80& 89.31 \\ 
        Qwen-2-72b & 81.68&89.93&88.47&81.90&87.48&85.70&89.54&88.12&83.72&87.73 \\ 
        Qwen-2.5-72b & 83.72&	87.88&	87.20&	80.68&	85.36&	83.89&	89.70&	88.75&	82.34&	87.42 \\ 
 Mistral-8x22b & 50.40& 57.08& 59.28& 44.02& 48.76& 58.94& 59.72& 62.44& 49.78& 57.83 \\ 
        Mistral-large-2 & 60.38& 76.08& 74.92& 60.19& 70.74& 68.50& 79.65& 78.84& 63.85& 71.66 \\ 
        Llama-3-70b & 65.17& 73.58& 75.22& 57.87& 72.84& 64.06& 72.82& 73.16& 57.34& 72.95 \\ 
        Llama-3.1-70b & 67.32& 76.57& 76.93& 60.96& 73.56& 72.23& 78.13& 78.23& 64.16& 74.90 \\ 
        Phi-3-medium & 45.26& 61.42& 58.40& 45.65& 51.33& 49.88& 59.33& 59.35& 45.49& 53.02 \\ 
        Gemma-2-27b & 48.50& 54.05& 53.32& 36.92& 48.22& 40.62& 41.72& 43.81& 32.99& 46.03 \\ 
        Yi-1.5-34b & 68.48& 81.92& 81.74& 70.89& 79.76& 74.13& 85.12& 83.38& 78.20& 80.30 \\ 
        Internlm-2.5-20b-chat & 67.16& 81.56& 77.72& 73.05& 72.64& 66.22& 82.65& 78.42& 72.94& 74.03 \\ 
        Internlm-2.5-20b & 72.86& 86.10& 82.14& 79.06& 74.70& 69.65& 78.79& 76.56& 70.28& 77.20 \\ 
        ERNIE-Lite & 53.45& 67.56& 67.73& 61.21& 61.21& 60.74& 70.27& 71.5& 62.43& 64.84 \\ 
        ERNIE-Tiny & 34.78& 37.86& 37.88& 33.08& 32.29& 32.52& 38.63& 37.58& 32.52& 34.6 \\ 
        ERNIE-turbo & 43.34& 56.05& 53.97& 52.02& 44.82& 41.01& 57.66& 54.28& 49.49& 46.95 \\ 
        Sensechat-5 & 69.97& 83.21& 80.73& 73.86& 76.95& 68.98& 82.00& 79.88& 73.52& 74.77 \\ 
        Claude-3.5 & 66.47& 76.84& 78.04& 60.60& 75.98& 75.92& 81.65& 84.24& 62.83& 82.54 \\ 
GLM-4 & 64.23& 84.39& 80.06& 75.66& 75.75& 72.18& 84.20& 80.07& 76.00& 78.06 \\ 
        ChatGPT & 49.78& 58.13& 58.74& 45.46& 52.42& 60.28& 59.81& 60.61& 47.50& 54.54 \\ 
        GPT-4o & 74.16& 83.28& 84.12& 71.60& 84.32& 72.35& 85.03& 84.32& 72.74& 81.58 \\ 
        GPT-4 & 67.68& 75.29& 77.26& 60.12& 74.46& 71.19& 76.75& 77.56& 63.50& 74.57 \\
        \textbf{YueTung-7b} & \textbf{93.01} & \textbf{92.54} & \textbf{89.84} & \textbf{90.81} & \textbf{91.55} & \textbf{93.36} & \textbf{93.27} & \textbf{91.04} & \textbf{91.77} & \textbf{91.85}\\
        
\bottomrule
\end{tabular}
\caption{\textbf{All results} of the comparison between texts generated by YueTung-7b and baselines in Yue-MMLU based on 0-shot and 5-shot settings and the correct texts.}
\label{MMLU_Cant_all}
\end{table*}

\subsubsection{Mainstream Language Benchmarks}

\begin{table*}
\centering
\begin{tabular}{l|c|c|c|c|c|c}
\toprule
\multirow{2}{4cm}{\textbf{Models \\ (English-TruthfulQA)}} & \multicolumn{3}{c|}{\textbf{0-shot (correct)}} & \multicolumn{3}{c}{\textbf{5-shot (correct)}} \\
\cmidrule{2-7}
& \textbf{Rouge-l} & \textbf{Bleu-4} & \textbf{BERTScore} & \textbf{Rouge-l} & \textbf{Bleu-4} & \textbf{BERTScore} \\
\midrule
Qwen-1.5-110b & 22.57 & 15.54 & 85.78 & 29.44 & 23.14 & 86.35 \\
Qwen-2-7b & 10.98 & 10.20 & 83.86 & 23.67 & 18.60 & 86.09 \\
Qwen-2-72b & 3.03 & 7.58 & 81.78 & 7.45 & 9.59 & 82.98 \\
Qwen-2.5-72b & 13.05 & 10.83  & 84.5 & 21.16 & 13.65  & 85.71 \\
Mistral-8x22b & 18.59 & 12.91 & 85.78 & 31.05 & 20.61 & 87.58 \\
Mistral-large-2 & 20.57 & 14.63 & 85.69 & 41.46 & 28.92 & 88.30 \\
Llama-3-8b & 16.89 & 11.59 & 84.11 & 58.34 & 38.35 & 88.50 \\
Llama-3-70b & 12.09 & 10.46 & 83.84 & 53.00 & 36.77 & 88.94 \\
Llama-3.1-8b & 14.13 & 11.34 & 83.46 & 51.70 & 36.95 & 88.47 \\
Llama-3.1-70b & 18.12 & 13.24 & 84.18 & 55.22 & 40.54 & 88.88 \\
Phi-3-medium & 27.90 & 17.35 & 86.48 & 43.02 & 28.62 & 88.24 \\
Gemma-2-27b & 12.31 & 9.84 & 83.56 & 18.25 & 12.25 & 84.31 \\
Yi-1.5-34b & 17.22 & 13.22 & 84.79 & 35.33 & 25.82 & 87.56 \\
Internlm-2-7b & 47.58 & 28.78  & 87.13 & 41.57 & 30.32  & 65.51 \\
Internlm-2-7b-chat & 9.54 & 9.69  & 83.42 & 23.39 & 18.97  & 86.29 \\
Internlm-2-20b & 43.50 & 27.33   & 87.5 & 41.13 & 31.64   & 85.39 \\
Internlm-2-20b-chat & 4.81 & 8.14   & 82.11 & 31.44 & 24.45   & 85.8 \\
Internlm-2.5-7b & 34.44 & 18.62 & 86.06 & 39.19 & 25.39 & 87.31 \\
Internlm-2.5-7b-chat & 7.45 & 8.82  & 82.92 & 12.92 & 11.29  & 84.39 \\
ChatGPT & 37.81 & 21.95 & 87.20 & 50.43 & 31.44 & 88.55 \\
GPT-4o & 17.93 & 13.05 & 85.38 & 49.52 & 37.44 & 88.62 \\
GPT-4 & 19.58 & 14.10 & 85.19 & 53.18 & 39.22 & 88.85 \\
\midrule
Qwen-2.5-7b& 9.46&11.70&82.95&15.47&10.93&84.33 \\
YueTung-7b& 37.41&22.54&89.15&63.50&36.13&93.14 \\
\bottomrule
\end{tabular}
\caption{Results of the comparison between texts generated by various LLMs in English-TruthfulQA based on 0-shot and 5-shot settings and the \textbf{correct} texts. \textbf{Rouge-l}, \textbf{Bleu-4}, and \textbf{BERTScore} are evaluation metrics for comparing text similarity.}
\label{TruthfulQA_Eng}
\end{table*}

\begin{table*}[h]
\centering
\begin{tabular}{l|c|c}
\toprule
\textbf{Models} & \textbf{Acc. (0-shot)} & \textbf{Acc. (5-shot)} \\
\midrule
Qwen-1.5-110b & 88.55 & 88.93 \\
Qwen-2-7b & 84.15 & 84.76 \\
Qwen-2-72b & 92.8 & 91.58 \\
Qwen-2.5-72b & 93.25 & 96.13  \\
Mistral-8x22b & 91.51 & 91.58 \\
Mistral-large-2 & 95.38 & 95.15 \\
Llama-3-8b & 80.36 & 81.05 \\
Llama-3-70b & 93.4 & 93.33 \\
Llama-3.1-8b & 85.97 & 86.35 \\
Llama-3.1-70b & 95.3 & 95.3 \\
Phi-3-medium & 90.3 & 90.83 \\
Gemma-2-27b & 24.49 & 9.86 \\
Yi-1.5-34b & 87.95 & 88.4 \\
Internlm-2-7b & 46.63 & 61.56  \\
Internlm-2-7b-chat & 73.54 & 66.64 \\
Internlm-2-20b & 78.54 & 64.14   \\
Internlm-2-20b-chat & 78.54 & 75.28  \\
Internlm-2.5-7b & 77.48 & 65.88  \\
Internlm-2.5-7b-chat & 84.99 & 82.71  \\
ChatGPT & 65.28 & 67.25 \\
GPT-4o & 95.22 & 95.68 \\
GPT-4 & 95.00 & 94.77 \\
\midrule
Qwen-2.5-7b& 88.62&88.65 \\
YueTung-7b& 84.32&86.26 \\
\bottomrule
\end{tabular}
\caption{Results of the comparison between answer generated by various LLMs in English-GSM8K based on 0-shot and 5-shot settings and groundtruth.}
\label{GSM8K_Eng}
\end{table*}

\begin{table*}[h]
\centering
\begin{tabular}{l|c|c}
\toprule
\textbf{Models} & \textbf{Acc. (0-shot)} & \textbf{Acc. (5-shot)} \\
\midrule
Qwen-1.5-110b & 82.66 & 77.6 \\ 
Qwen-2-7b & 65.41 & 69.7 \\
Qwen-2-72b & 69.79 & 79.83 \\ 
Qwen-2.5-72b & 95.19 & 94.76 \\ 
        Mistral-8x22b & 90.82 & 88.07 \\ 
        Mistral-large-2 & 94.51 & 94.59 \\ 
Llama-3-8b & 81.63 & 78.88 \\
        Llama-3-70b & 93.22 & 92.62 \\ 
Llama-3.1-8b & 80.52 & 84.21 \\
        Llama-3.1-70b & 93.56 & 93.3 \\ 
Phi-3-medium & 93.13 & 92.1 \\
Gemma-2-27b & 82.92 & 72.79 \\
Yi-1.5-34b & 92.36 & 92.53 \\
Internlm-2.5-7b & 85.58 & 85.15 \\
Internlm-2.5-7b-chat & 87.04 & 86.78 \\
\midrule
Qwen-2.5-7b&83.95 &78.20 \\
YueTung-7b& 89.15&95.25 \\
\bottomrule
\end{tabular}
\caption{Results of the comparison between answer generated by various LLMs in English-ARC challenge based on 0-shot and 5-shot settings and groundtruth.}
\label{ARC_Eng}
\end{table*}

\begin{table*}\small
\centering
\begin{tabular}{l|c|c|c|c|c|c|c|c|c|c}
\toprule
\multirow{2}{4cm}{\textbf{Models \\ (Standard Chinese-MMLU)}} & \multicolumn{5}{c|}{\textbf{0-shot (correct)}} & \multicolumn{5}{c}{\textbf{5-shot (correct)}} \\
\cmidrule{2-11}
& \textbf{STEM} & \textbf{Hum.} & \textbf{S.S.} & \textbf{C.S.} & \textbf{Oth.} & \textbf{STEM} & \textbf{Hum.} & \textbf{S.S.} & \textbf{C.S.} & \textbf{Oth.}\\
\midrule
 Qwen-1.5-110b& 78.06 & 87.6 & 85.88 & 81.83 & 84.04 & 85.1 & 90.77 & 91.07 & 85.84 & 91.56 \\ 
Qwen-2-7b& 77.52 & 86.63 & 85.1 & 77.37 & 83.41 & 81.62 & 86.94 & 85.09 & 80.06 & 83.84 \\
        Qwen-2-72b& 83.36 & 89.69 & 88.75 & 83.16 & 86.58 & 90.07 & 93.18 & 92.97 & 88.64 & 91.07 \\ 
        Qwen-2.5-72b& 83.26 & 89.54 & 89.14 & 82.04 & 88.33 & 85.87 & 90.6 & 90.25 & 84.15 & 88.4 \\ 
        Mistral-8x22b & 57.88 & 63.27 & 64.51 & 49.18 & 57.28 & 62.38 & 62.97 & 63.7 & 51.52 & 58.26 \\ 
        Mistral-large-2 & 68.49 & 79.48 & 77.03 & 64.36 & 70.8 & 71.65 & 81.95 & 78.76 & 66.87 & 74.52 \\ 
        Llama-3-8b& 54.04 & 61.35 & 59.17 & 45.67 & 56.28 & 47.66 & 59.26 & 58 & 44.72 & 53.54 \\
        Llama-3-70b & 72.64 & 77.23 & 77.44 & 60.22 & 76.3 & 72.04 & 75.31 & 74.99 & 58.74 & 74.72 \\ 
        Llama-3.1-8b& 49.08 & 61.05 & 59.17 & 44.15 & 53.11 & 55.62 & 62.58 & 61.02 & 46.43 & 56.27 \\
        llama-3.1-70b & 69.84 & 77.77 & 76.9 & 62.34 & 75.02 & 72.4 & 77.95 & 78.57 & 61.6 & 75.75 \\ 
        Phi-3-medium& 58.54 & 63.46 & 65.61 & 48.45 & 61.5 & 57.18 & 62.84 & 66.32 & 49.76 & 59.06 \\
        Gemma2-27b& 49.67 & 53.63 & 57.23 & 42.36 & 50.35 & 40.25 & 43.15 & 47.77 & 37.14 & 46.34 \\
        Yi-1.5-34b& 73.02 & 83.78 & 82.99 & 74.6 & 83.72 & 78.87 & 86.24 & 84.47 & 77.68 & 85.06 \\
        Internlm-2.5-7b& 75.62 & 88 & 83.95 & 79.14 & 80.86 & 70.52 & 87.27 & 83.38 & 79.6 & 80.19 \\
        Internlm-2.5-7b-chat& 73.04 & 87.42 & 84.23 & 77.62 & 85.29 & 69.24 & 86.45 & 83.78 & 77.93 & 83.46 \\
\midrule
Qwen-2.5-7b&71.29 &82.50&78.43&69.57&74.04&75.77&84.26&81.64&71.72&78.54 \\
YueTung-7b& 94.08&92.81&90.68&91.92&92.63&94.48&93.39&91.98&93.57&94.49 \\
        
\bottomrule
\end{tabular}
\caption{Results of the comparison between texts generated by various LLMs in CMMLU based on 0-shot and 5-shot settings and the correct texts.}
\label{CMMLU}
\end{table*}

\end{document}